\pdfoutput=1
\documentclass[conference]{IEEEtran}
\IEEEoverridecommandlockouts
\usepackage{cite}
\usepackage{amsmath,amssymb,amsfonts}
\usepackage{algorithmic}
\usepackage{graphicx}
\usepackage{textcomp}
\usepackage{xcolor}

\usepackage{array}
\newcolumntype{_}{>{\global\let\currentrowstyle\relax}}
\newcolumntype{^}{>{\currentrowstyle}}
\newcommand{\rowstyle}[1]{\gdef\currentrowstyle{#1}%
  #1\ignorespaces
}

\usepackage{multirow}
\usepackage{colortbl}
\usepackage{svg}
\usepackage{ctable}

\def\BibTeX{{\rm B\kern-.05em{\sc i\kern-.025em b}\kern-.08em
    T\kern-.1667em\lower.7ex\hbox{E}\kern-.125emX}}
\begin{document}

\title{Generalization in Multimodal Language Learning from Simulation
}

\author{\IEEEauthorblockN{Aaron Eisermann, Jae Hee Lee, Cornelius Weber, Stefan Wermter}
\IEEEauthorblockA{\textit{Knowledge Technology, Department of Informatics} \\
\textit{University of Hamburg}\\
Hamburg, Germany \\
aaron.eisermann@studium.uni-hamburg.de, \{lee, weber, wermter\}@informatik.uni-hamburg.de}
}

\makeatletter
\def\ps@IEEEtitlepagestyle{%
  \def\@oddfoot{\mycopyrightnotice}%
  \def\@oddhead{\hbox{}\@IEEEheaderstyle\leftmark\hfil\thepage}\relax
  \def\@evenhead{\@IEEEheaderstyle\thepage\hfil\leftmark\hbox{}}\relax
  \def\@evenfoot{}%
}
\def\mycopyrightnotice{%
  \begin{minipage}{\textwidth}
  \centering
  \copyright~2021 IEEE. Personal use of this material is permitted.
  \end{minipage}
}
\makeatother

\maketitle

\begin{abstract}

Neural networks can be powerful function approximators, which are able to model high-dimensional feature distributions from a subset of examples drawn from the target distribution. Naturally, they perform well at generalizing within the limits of their target function, but they often fail to generalize outside of the explicitly learned feature space. It is therefore an open research topic whether and how neural network-based architectures can be deployed for systematic reasoning. Many studies have shown evidence for poor generalization, but they often work with abstract data or are limited to single-channel input. Humans, however, learn and interact through a combination of multiple sensory modalities, and rarely rely on just one. To investigate compositional generalization in a multimodal setting, we generate an extensible dataset with multimodal input sequences from simulation. We investigate the influence of the underlying training data distribution on compostional generalization in a minimal LSTM-based network trained in a supervised, time continuous setting. We find compositional generalization to fail in simple setups while improving with the number of objects, actions, and particularly with a lot of color overlaps between objects. Furthermore, multimodality strongly improves compositional generalization in settings where a pure vision model struggles to generalize.

\end{abstract}

\begin{IEEEkeywords}
Robotics, Computer vision, Language generation
\end{IEEEkeywords}

\section{Introduction}

In the last decade neural network-based algorithms have become a popular method for machine learning tasks, partially fueled by the hardware advancements of GPUs to effectively train large neural network models. As a result, various architectures have been used to build powerful artificial intelligent agents that rival or even outperform human capabilities on some well-defined tasks. One of the largest benefactors might be computer vision-based tasks, where models based on convolutional layers quickly became state-of-the-art \cite{krizhevsky2012imagenet}, proving their applicability for general object classification. For tasks involving sequence-to-sequence transduction, recurrent neural network models have been developed that can deal with time continuous data, which is useful in tasks like natural language processing.

The common goal behind neural networks is to approximate a function that maps high-dimensional and complex inputs to a target distribution by both inter- and extrapolating commonalities from examples that are shown during training. They are expected to generalize well within the close definition of the target distribution, but tend not to reason outside of it \cite{loula2018rearranging,lake2017generalization}. The task of \textit{compositional generalization}, which requires combining known elements in novel ways, is challenging for neural networks to tackle. Compositional generalization tasks can be formulated at arbitrary levels of difficulty, but fail often in even the simplest settings. Ruis et al.~\cite{ruis2020benchmark} demonstrate this in multiple experiments in an abstract grid world. In their experiments, they trained an agent in the grid world commands such as pushing a certain-colored shape, and then asked the model to recompose to novel combinations. Their model was unable to put objects into a context it has not seen before, even if they appeared in the training data with partial labeling.

\begin{figure}
  \centering%
  \includegraphics[trim=15 20 10 1, clip,width=\linewidth]{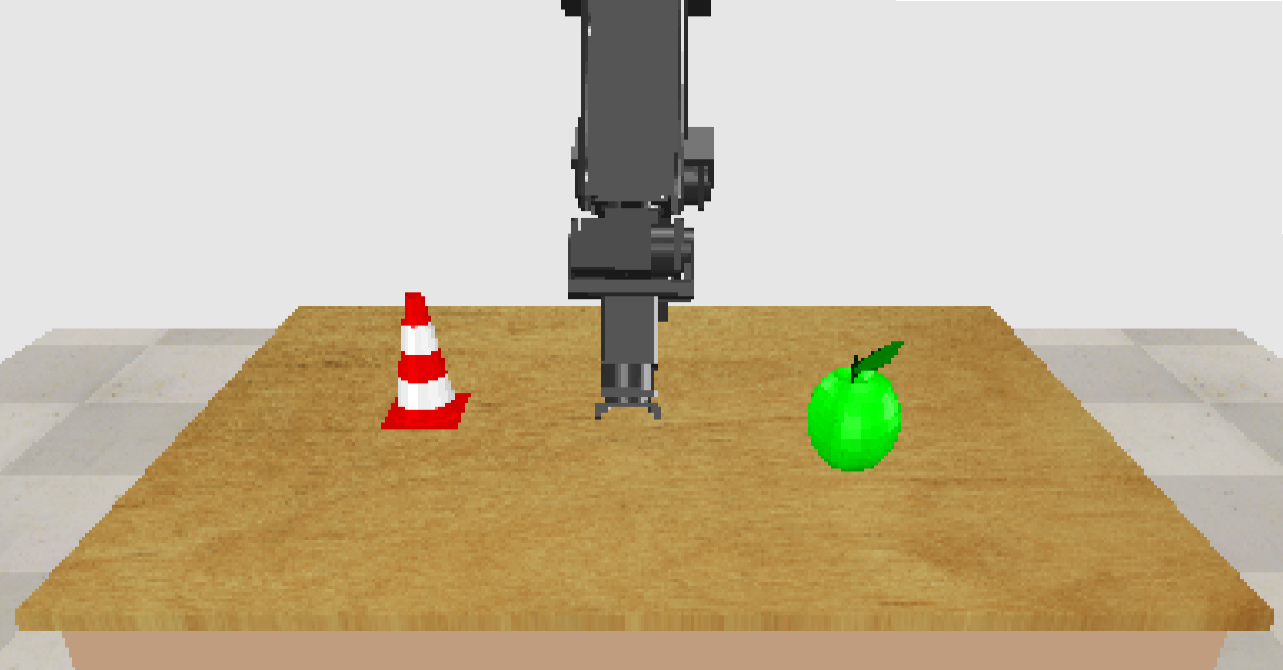}
  \caption{The view of an example scene. The manipulator is in its initial pose. In the shown setting, two interactable objects are placed randomly on the table.}
  \label{fig:CoppeliaSimVisionSensor}
\end{figure}
  
\begin{figure*}
  \centering%
  \includegraphics[width=\linewidth]{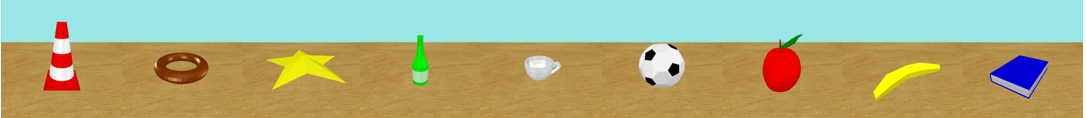}
  \caption{The nine objects used for generating the dataset.}
  \label{fig:CoppeliaSimObjects}
\end{figure*}
Although neural networks are machine learning tools that are generally applicable, the individual models are often tailored and limited to one clearly defined input, such as a camera image for object detection, an audio signal for language recognition, or written text for machine translation. Humans, on the other hand, receive many sensory inputs concurrently, such as additional tactile and sensorimotor feedback signals, from which they are able to identify the perceptions that are useful in a given context, focus on them and disregard unimportant information. This exposure to multiple modalities shapes their way of learning early on, as children learn best through interacting with their environment and experiencing many different sensations. It enables them to learn the complex rules of a language passively within few years and apply these rules to words they hear for the first time.
Roboticists advocate grounded language learning where an artificial agent actively interacts with its environment sampling multimodal stimuli while learning language \cite{steels2012language,cangelosi2010139,cohen2018194}.

This \textit{multimodality} could be a strong driver of systematic generalization, by teaching us to mix and match perceptions and experiences. Heinrich et al.~\cite{heinrich_2020} apply multimodal integration to neural networks to simulate the development of intelligent agents that can learn from experiences in a human-like manner, based on the small but real multimodal robotic dataset EMIL \cite{heinrich_2018_EMIL}. Inspired by their work, we take the question of compositional generalization in grounded language learning into a 3D simulation framework, as shown in Figure~\ref{fig:CoppeliaSimVisionSensor}. We generate larger, simulated multimodal datasets to search for characteristics in the richness of the training data that enable generalization. Based on a single, relatively simple recurrent neural architecture, which we train with the various datasets, and using the same generalization test set for all cases, we show how compositional generalization can emerge by providing more varying but related training data.

\section{Related Work}

There are various ways of measuring compositional generalization, and the underlying data strongly influences what can be tested. The generation of meaningful benchmarks for compositional generalization is a topic of current research, and multiple proposals have been made to this end.
In the experiments of Ruis et al.~\cite{ruis2020benchmark}, the GroundedSCAN benchmark was used for grounded language learning in an abstract grid world. A tensor describes the exact state of the grid world via one-hot encoding. For each grid cell that contains an object, the three properties shape, color, and size are given. The agent's position and orientation are also encoded. For their experiments, Ruis et al.\ developed a context-free grammar to generate commands for the agent to follow, incorporating also sentences that require a contextual understanding of words like ``small'' when the same object is present in different sizes. They find that in most of their tests on compositional generalization their models fail.
Shaw et al.\ \cite{shaw2020compositional} claim that the SCAN benchmark does not correlate well with non-synthetic data, and argue that most research on compositional generalization focuses on specialized architectures that introduce strong compositional biases. They apply a systematic method of generating a compositional train-test split by maximizing the difference between word compounds while minimizing the difference of the distribution of atoms~\cite{keysers2019measuring}. Yet these datasets disregard the continuous nature of space.

CLEVR \cite{johnson2017clevr} is a dataset for visual reasoning, where the agent has to answer questions based on a rendered input image with simple 3D shapes that assume different colors and materials. The authors discover that models learn strong biases about the color of objects. They also find that their model composes to unseen shape-material combinations more readily than shape-color combinations. CLEVR, however, disregards any agent interactions with the objects.


This paper is inspired by the the work of \cite{heinrich_2020}, where the authors evaluate multiple continuous time recurrent neural network architectures to learn a language representation that is grounded in the auditory, sensorimotor, and visual perceptions of a robot, obtained from the EMIL dataset \cite{heinrich_2018_EMIL}. The EMIL dataset contains recordings of a humanoid robot interacting with objects on a table in a child-like manner, pushing and grasping the objects. An independent teacher provides descriptions of the interactions of a robot in natural language. They report that all models struggle on generalization tasks. They also find that their models tend to rely on a single modality when the training dataset is smaller, performing better without the other modalities. With a larger training set however, the model benefits from the additional modalities.

Zhong et al.~\cite{zhong2019sensorimotor} use a similar dataset to train a model to execute spoken commands provided as action-object pairs. Speech recognition software is used to translate the spoken command into discrete values defined in a lookup table. The authors highlight the use of a large robotic dataset, which was deemed more difficult than generalization on simpler datasets~\cite{zhong2019sensorimotor}.

In the research area of robotics, unsupervised approaches are well-adopted approaches, most importantly reinforcement learning (RL), because the creation of useful supervised training data is often not feasible. RL with real robots however is costly and time-consuming, so many approaches are trained and evaluated in simulation before being transferred to the real world \cite{james2019sim,peng2018sim}. Especially robots that interact with the real world need to be able to systematically generalize to new situations.
Results in simulation \cite{hill2019environmental} suggest that the additional complexity of 3D environments helps in learning systematical generalization. It was found that an egocentric view, which forces an agent to experience the scene from different angles, improves generalization performance, compared to a global view that is typical for 2D experiments. Furthermore, a task of object detection, where a classifier and a reinforcement learning agent had to classify between two options, was solved better by the dynamic agent than by a static classifier system.

\section{Data for Grounded Language Learning}

\begin{figure}
  \includegraphics[width=\linewidth]{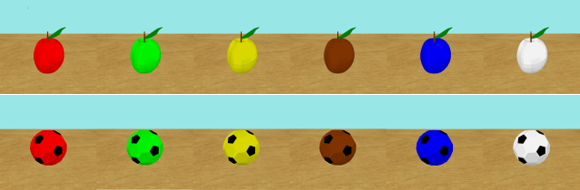}
  \caption{Two objects (an apple and a soccer ball) are shown in all six colors used in the dataset.}
  \label{fig:CoppeliaSimColors}
\end{figure}

We set up a simplistic environment within the CoppeliaSim\footnote{https://www.coppeliarobotics.com/} robot simulator by Coppelia Robotics, where we placed a 3D-model of the OpenManipulator Pro\footnote{https://emanual.robotis.com/docs/en/platform/openmanipulator\_p/overview/} by Robotis on a table that confines a ``play area''. Figure \ref{fig:CoppeliaSimVisionSensor} shows a green apple and a red pylon placed randomly on the table so the manipulator can move them around. We also modeled nine different objects with low complexity but noticeable differences between their shapes, shown in Figure \ref{fig:CoppeliaSimObjects}. Any color can be painted onto the objects, although some of them have certain textural features that are exempt from changes of their color, e.g. black patches on a soccer ball (cf.~Figure \ref{fig:CoppeliaSimColors}).

The interactions we simulate are hard-coded: an action among the ones listed in Table~\ref{tab:actions} is performed by moving the robot arm according to the object's position. This is done by defining waypoints for the arm's endpoint trajectory and solving the inverse kinematics using the Damped Least Squares (DLS) method \cite{Wampler1986} with a dampening factor of 0.5 and a limit of 99 calculation iterations \cite{DeoWalker1995}. Due to the random placement of objects on the 2D table surface, an unlimited amount of variations exists for any action in principle. 
An \textit{interaction} is described as a combination of an \textit{action}, a \textit{color} adjective and an \textit{object}, such as picking up a green apple or pushing a blue banana to the left.

By recording continuous interactions we create data for the model's three input modalities that represent visual, sensorimotor and auditory sensations:

\begin{itemize}
\item \emph{Visual data} is obtained by a camera that is placed in front of the table, recording at a slightly downwards tilted angle.
\item \emph{Sensorimotor data} is gathered from the position readings of the robot arm's joints.
\item \emph{Language information} is provided as a descriptive commentary following an action. We use word-level one-hot encoding to form sentences structured as verb--color--noun over the course of multiple frames.
\end{itemize}
The data was captured at a framerate of 5 frames per second and the image size set to $398 \times 224$ pixels. Joint positions were the angles of the 6 degree-of-freedom robot arm in degrees. Words were uttered over the span of 0.5 seconds each, consequently spanning sometimes 2 or 3 frames at the recording framerate. There were 5 action words, 6 color words, 9 object words, and one ``n/a'' output dimension that signifies that none of the words is produced. This sums up to 21 language input as well as output units of which one is supposed to be activated at any time step. The left side of Figure~\ref{fig:model_general} illustrates the sequence of inputs.


\section{Model}

At training time, we want our model to continuously receive inputs from the three modalities, out of which the auditory or language input is a hindsight description of an interaction. This is inspired by the idea of an external teacher, like a parent, who provides explanations to a child interacting with its environment. Eventually, the model, like a child, should be able to generate these descriptions by itself and learn to understand the underlying rules the language abides by.

\begin{figure*}
    \centering
    \includegraphics[trim=607 392 220 135, clip,width=\linewidth]{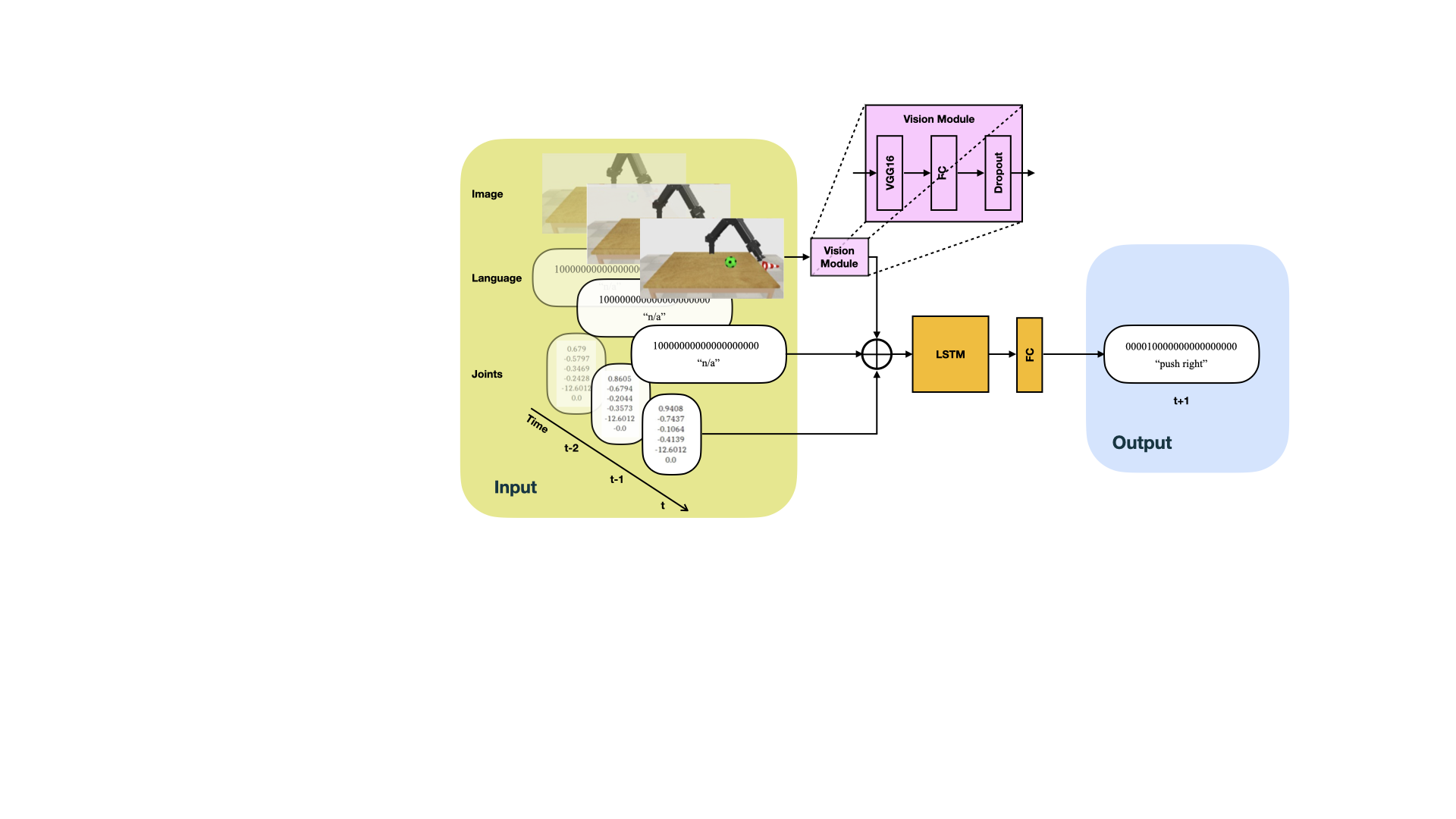}
    \caption{Network architecture. The network reads sequences of image, language and joints inputs. The vision module processes the image by a VGG-16 CNN and reduces dimensionality to 32 neurons by a fully-connected layer. A dropout layer deactivates $50\%$ of the neurons during training. The resulting output is concatenated with the 6-dimensional joint and 21-dimensional language vectors and fed into a single-layer LSTM with 256 hidden neurons. The output from the LSTM is mapped via a fully-connected layer to the 21-dimensional output layer.}
    \label{fig:model_general}
\end{figure*}

Figure \ref{fig:model_general} shows how the three inputs are handled by the model to produce the output. Compared to the joint and language data, which can be represented as one-dimensional arrays of sizes 6 and 21 respectively, the image data is  high-dimensional. 
We therefore use a size reduction of the image features, which are obtained from the pre-trained VGG-16 network~\cite{simonyan2014very} (as visual features we use the output before the first fully-connected layer of VGG-16), by feeding the image features to a fully-connected layer. A dropout layer is used for regularization.

We used curriculum learning to encourage the models to learn words more independently, which also led them to learn faster~\cite{bengio2009}. Thereby, a model learned to produce just one, two or all three words, dependent on a mask. To this end, we added three binary input neurons, which we refer to as ``mask neurons'', each corresponding to either the noun, adjective or verb in the sentence. The combination of these three neurons defines a masking configuration. Deactivating a mask neuron signals the model to exclude this part of the sentence by replacing the word with ``n/a''. This masking is applied to both the input word vector as well as the target output vector. During training a new masking configuration is generated randomly at the beginning of every sentence and the same mask is applied to the word vectors of all frames of the same sentence.

The masked word vector and the mask neurons, the joint vector, and the vision module output are concatenated and fed into a single layer LSTM. The LSTM module receives its previous hidden output to predict a new hidden output and make a language prediction via a subsequent fully-connected layer that reduces its output to word vector size. We use cross entropy loss for training the model. 

\begin{table}
\centering
\caption{The four actions of the robot arm used in the dataset.}
\label{tab:actions}
\begin{tabular}{|l|} 
\hline
``push right''\\
``push left''\\ 
``pick up''\\
``put down''\\
\hline
 \end{tabular}
\end{table}

\section{Evaluation}

We ran multiple experiments in which we gradually changed the composition of the training data to investigate the effects on compositional generalization.

\subsection{Data Generation Setup}
The data composition is controlled via four parameters during the data generation process. The first parameter controls the actions that are used. We change these between using only the two pushing actions and using all four actions. The second and third parameters control the number of objects and colors that are included. Lastly, we vary the number of objects that are visible at the same time to be either one or two.

In all experiments, we use the same ground rules to generate training and validation datasets, so they are drawn from the same feature space. The training sets always show 5000 interactions, while the validation sets include 2500 interactions. We also balance all datasets with respect to the possible combinations of objects, colors, and actions given the respective rule set, so that within one dataset all combinations occur with approximately the same count. Due to the constant size of all training sets, datasets featuring a greater variety of categories contain fewer samples per category.

\subsection{Training Setup}
We evaluate our model using the PyTorch deep learning library~\cite{NEURIPS2019_9015} on a machine equipped with 32 CPU cores (Intel Xeon Silver 4110 CPU@2.10GHz), 188GB RAM, and eight GPU cores (GeForce GTX 1080TI) with 11GB VRAM.

 We use ADAM~\cite{kingma2014adam} for optimization, where we reduce the learning rate by a factor of 0.1 starting from 0.001 if the validation performance does not improve for five epochs.

The probabilities we use to select a mask for each sentence were 15\% for one active neuron, 45\% for two active neurons, and 40\% for all neurons being active. 

\subsection{Metric}
Our use of cross entropy loss to train the model intuitively suggests a frame-wise evaluation to measure the performance of the model by averaging the accuracy of the prediction over all frames. This \textit{frame-wise accuracy} is a useful indicator of the training progress, as the model itself is being optimized on a frame-level basis. However, this measure does not resonate well with the higher goal of learning entire sentences. It is also an inflated measure, since most of the frames in our datasets are ``silent'' frames where the corresponding word is ``n/a''. Additionally, this measure is susceptible to small time shifts of the predicted output against the ground truth. These issues combined mean that, with the frame-wise accuracy, a model that always predicts ``n/a'' outscores a model that might predict the correct sentence to describe an interaction, but does so a few frames too early. We therefore calculate another measure of performance, the \textit{sentence-wise accuracy}, which compares the predicted sentence produced over all frames of an interaction to the true sentence. As we mainly use the sentence-wise accuracy for interpretation of our results, we refer to it as \textit{accuracy}.

\subsection{Test Sets}

For each experiment, we track the \textit{training} error on the corresponding training set and the \textit{validation} error on separate validation data from the same corresponding distribution.

Since the different experiments present different data, we also generate one \textit{constant test set}, which consists of such data that is common to all experiments.
We chose four interactions that make up the constant test set, namely ``push right white football'', ``push right yellow banana'', ``push left brown bottle'', and ``push left red ring''. This constant test set only includes these four interactions and they appear in the training set of all our experiments (with the same language outputs but with different images and joint values).

The \textit{compositional generalization test set} is generated from the actions in the opposite directions, thus it includes the interactions ``push \textit{left} white football'', ``push \textit{left} yellow banana'', ``push \textit{right} brown bottle'', and ``push \textit{right} red ring'', highlighted once more in Table~\ref{tab:leave-out-data}. The interactions in this set are the leave-out cases, which we use to test for compositional generalization. They are not included in the training data of any of our experiments.

\begin{table}
\centering
\caption{The leave-out data never shown during training for test of compositional generalization.}
\label{tab:leave-out-data}
\begin{tabular}{|l|} 
\hline
 ``push left white football''\\
 ``push left yellow banana''\\
 ``push right brown bottle''\\
 ``push right red ring''\\
\hline
 \end{tabular}
\end{table}

With these constraints, we generate one constant test set and one compositional generalization test set for the experiments in which we show only one object at a time. We generate another constant test set and compositional generalization test set for all experiments in which we show two objects at the same time. This adds up to four test sets in total. Each of the test sets contains 2000 interactions, which are 500 examples of each of the aforementioned interactions.

The variables that determine the different learning conditions are summarized as follows:
\begin{description}
\item[\textbf{A2} / \textbf{A4}] \hfill\\
  Two or four different actions exist in each scene.
\item[\textbf{V1} / \textbf{V2}] \hfill\\
  One or two objects are visible simultaneously in each scene.
\item[\textbf{O4} / \textbf{O9}] \hfill\\
  Four or nine object classes exist.
\item[\textbf{C1} / \textbf{C6}] \hfill\\
  An object can have one color, or differing colors from a selection of 6 colors in the dataset.
\item[\textbf{X}] \hfill\\
  The object colors in compositional generalization test set are exclusive, i.e., the color of an object in the compositiontal generalization test set is not shared with any another objects in the training set.
\item[\textbf{$\neg$X}] \hfill\\
  The object colors in the compositional generalization test set have an overlap with the object colors in the training set, i.e., there exists at least one other object of same color in the training set (e.g., while the network has never seen during trainining ``push right red ring'', which is in the compositional generalization test set, it might have seen ``push right red star'').
\item[\textbf{Without Joint Readings}] \hfill\\
  The network does not receive the joint vector as input and must rely solely on vision to describe the scene.
\end{description}

\newcommand{\rotaterow}[4]{\multirow{#1}{*}{\rotatebox{90}{\parbox{#2}{\centering #3\\ #4}}}}

\subsection{Results}

We gathered results by repeating each experimental setup three times, averaging the results. We report mean and standard deviation of the sentence-wise accuracy of the respective training and validation sets as well as the two standardized test sets. The reported training and validation accuracy were computed post-training by evaluating the final model on the training and validation sets. Table~\ref{tab:results_sentence_all} summarizes the results. 
\begin{table*}[tb]
  \centering
  \caption{Mean sentence-wise accuracy in percent of the models trained in various conditions; parentheses denote standard deviation}
  \label{tab:results_sentence_all}
  \begin{tabular}{_r^c^l^c^c^c^c^c^c}
    \toprule
                                                           &                                               &                              & \multicolumn{2}{c}{4 Objects (\textbf{O4})}       & \multicolumn{4}{c}{9 Objects (\textbf{O9})}                                                                                                                              \\
                                                           &                                               & \multicolumn{1}{c}{}         & \multicolumn{2}{c}{Exclusive Colors (\textbf{X})} & \multicolumn{2}{c}{Color Overlap (\textbf{$\neg$X})} & \multicolumn{2}{c}{Exclusive Colors (\textbf{X})}                                                                 \\
    \cmidrule(lr){4-5}\cmidrule(lr){6-7}\cmidrule(lr){8-9}
                                                           &                                               &                              & 2 Actions (\textbf{A2})                           & 4 Actions (\textbf{A4})                              & 2 Actions (\textbf{A2}) & 4 Actions (\textbf{A4}) & 2 Actions (\textbf{A2}) & 4 Actions (\textbf{A4})             \\
    \midrule[\heavyrulewidth]
    \rotaterow{8}{2.8cm}{1 Visible Object}{(\textbf{V1})}  & \rotaterow{4}{1.3cm}{1 Color}{(\textbf{C1})}
                                                           & Training                                      & 95.59 (1.8)                  & 95.33 (1.53)                                      & 94.03 (0.39)                                         & 90.32 (3.09)            & 94.63 (1.3)             & 91.5 (1.29)                                                   \\
                                                           &                                               & Validation                   & 95.8 (1.64)                                       & 94.07 (1.56)                                         & 91.07 (0.45)            & 87.51 (2.78)            & 94.21 (1.28)            & 89.93 (1.39)                        \\
                                                           &                                               & Constant Test Set            & 94.97 (2.01)                                      & 84.77 (3.69)                                         & 94.25 (0.04)            & 85.23 (1.61)            & 95.27 (0.78)            & 81.57 (1.02)                        \\
                                                           &                                               & \textbf{Comp.~Gen.~Test Set} & \rowstyle{\bfseries}0.0 (0.0)                     & 0.0 (0.0)                                            & 37.48 (3.6)             & 48.6 (7.07)             & 17.97 (4.34)            & 26.63 (3.91) \rowstyle{\normalfont} \\
    \cmidrule(l){2-9}
                                                           & \rotaterow{4}{1.3cm}{6 Colors}{(\textbf{C6})} & Training                     & 87.26 (1.15)                                      & 86.67 (1.61)                                         & 76.87 (3.02)            & 84.19 (1.59)            & \multicolumn{2}{c}{\multirow{4}{*}{N/A}}                      \\
                                                           &                                               & Validation                   & 84.85 (0.61)                                      & 83.52 (1.05)                                         & 73.83 (3.48)            & 76.71 (1.38)            &                         &                                     \\
                                                           &                                               & Constant Test Set            & 84.48 (1.55)                                      & 80.65 (2.65)                                         & 76.45 (4.21)            & 72.25 (1.35)            &                         &                                     \\
                                                           &                                               & \textbf{Comp.~Gen.~Test Set} & \rowstyle{\bfseries}63.4 (6.22)                   & 66.1 (3.0)                                           & 67.48 (2.78)            & 65.62 (3.58)            &                         & \rowstyle{\normalfont}              \\
    \cmidrule(l){1-9}
    \rotaterow{8}{2.8cm}{2 Visible Objects}{(\textbf{V2})} & \rotaterow{4}{1.3cm}{1 Color}{(\textbf{C1})}  & Training                     & 95.65 (0.4)                                       & 95.75 (0.62)                                         & 78.23 (1.63)            & 84.84 (2.65)            & 81.49 (0.81)            & 87.11 (0.81)                        \\
                                                           &                                               & Validation                   & 94.69 (0.59)                                      & 93.2 (0.63)                                          & 72.89 (1.38)            & 77.21 (2.15)            & 78.6 (0.59)             & 78.33 (1.11)                        \\
                                                           &                                               & Constant Test Set            & 95.28 (0.79)                                      & 87.9 (0.54)                                          & 81.9 (0.47)             & 75.27 (3.7)             & 85.1 (1.34)             & 76.7 (0.32)                         \\
                                                           &                                               & \textbf{Comp.~Gen.~Test Set} & \rowstyle{\bfseries}0.0 (0.0)                     & 0.0 (0.0)                                            & 0.1 (0.14)              & 0.72 (0.29)             & 0.02 (0.02)             & 1.25 (0.99) \rowstyle{\normalfont}  \\
    \cmidrule(l){2-9}
                                                           & \rotaterow{4}{1.3cm}{6 Colors}{(\textbf{C6})} & Training                     & 47.04 (0.51)                                      & 54.84 (2.73)                                         & 39.69 (1.2)             & 45.93 (2.1)             & \multicolumn{2}{c}{\multirow{4}{*}{N/A}}                      \\
                                                           &                                               & Validation                   & 40.92 (0.92)                                      & 42.49 (1.79)                                         & 34.0 (1.44)             & 37.27 (0.92)            &                         &                                     \\
                                                           &                                               & Constant Test Set            & 39.47 (1.36)                                      & 33.12 (0.44)                                         & 32.78 (2.28)            & 28.1 (0.05)             &                         &                                     \\
                                                           &                                               & \textbf{Comp.~Gen.~Test Set} & \rowstyle{\bfseries}9.55 (1.99)                   & 10.28 (1.34)                                         & 16.22 (1.9)             & 15.75 (1.95)            &                         & \rowstyle{\normalfont}              \\    
    \midrule[\heavyrulewidth]
    \multicolumn{9}{c}{\textbf{Without Joint Readings}}                                                                                                                                                                                                                                                                                                                           \\
    \midrule
    \rotaterow{8}{2.8cm}{1 Visible Object}{(\textbf{V1})} & \rotaterow{4}{1.3cm}{1 Color}{(\textbf{C1})}  & Training                     & 89.04 (3.72)                     & 58.8 (8.99)   & 77.03 (1.35)  & 27.65 (4.78) & 76.2 (5.88)  & 40.49 (6.44)                                                                                                                 \\
                                                          &                                               & Validation                   & 88.49 (3.45)                     & 56.37 (8.27)  & 75.28 (1.56)  & 27.69 (4.81) & 74.08 (6.42) & 38.73 (5.81)                                                                                                                 \\
                                                          &                                               & Constant Test Set            & 88.58 (3.65)                     & 68.23 (6.86)  & 79.1 (1.72)   & 51.9 (2.47)  & 76.78 (3.93) & 63.15 (4.39)                                                                                                                 \\
                                                          &                                               & \textbf{Comp.~Gen.~Test Set} & \rowstyle{\bfseries}0.0 (0.0)    & 0.0 (0.0)     & 12.05 (6.37)  & 0.15 (0.11)  & 5.12 (0.91)  & 0.13 (0.08) \rowstyle{\normalfont}                                                                                           \\
    \cmidrule(l){2-9}
                                                          & \rotaterow{4}{1.3cm}{6 Colors}{(\textbf{C6})} & Training                     & 76.19 (1.3)                      & 26.71 (8.9)   & 61.47 (7.84)  & 27.59 (3.67) & \multicolumn{2}{c}{\multirow{4}{*}{N/A}}                                                                                                    \\
                                                          &                                               & Validation                   & 72.83 (1.08)                     & 25.41 (8.55)  & 60.25 (8.4)   & 24.48 (2.85) &              &                                                                                                                              \\
                                                          &                                               & Constant Test Set            & 74.75 (1.05)                     & 27.92 (10.97) & 61.12 (7.09)  & 28.48 (4.95) &              &                                                                                                                              \\
                                                          &                                               & \textbf{Comp.~Gen.~Test Set} & \rowstyle{\bfseries}47.07 (2.07) & 6.67 (2.75)   & 50.28 (10.62) & 19.33 (3.56) &              & \rowstyle{\normalfont}                                                                                                       \\
    
    \bottomrule
  \end{tabular}
\end{table*}
The main findings are
\begin{enumerate}
\item The simplest setting led to lowest, 0\%, generalization, despite highest training and validation accuracies around 95\%.

\item Providing additional objects for training increased compositional generalization accuracies. 

\item Color overlap between different objects increased compositional generalization accuracies.

\item Showing two objects in the scene simultaneously, i.e. the object that is acted upon plus a distractor object, disrupted performance significantly.

\item Removing a sensory modality, i.e.~leaving away the joint readings, reduced training, validation and generalization performance.
\end{enumerate}
We detail the main findings in the following subsections.

\subsubsection{The simplest setting led to lowest, 0\%, generalization, despite highest training and validation accuracies around 95\%} This is shown in the top-left segment of Table~\ref{tab:results_sentence_all}, where 4 objects of a unique color each (C1) were shown and the agent was performing either 2 actions (A2) or 4 actions (A4). Note that in the case of 2 actions, the networks have been trained only with the 4 classes of data of the constant test set, i.e. similar to Table~\ref{tab:leave-out-data} but with action into the opposite directions, while in the case of 4 actions, the networks have seen 12 different classes of actions during training.

\subsubsection{Providing additional objects for training increased compositional generalization accuracies} This can be seen by comparing the two left-most columns (O4) with the four right-most columns (O9) in Table~\ref{tab:results_sentence_all}. The finding can be seen across experimental conditions. While improving compositional generalization, the larger task complexity when having more objects led in turn to decreased accuracies on the training, validation and constant test sets.

\subsubsection{Color overlap between different objects increased compositional generalization accuracies} This can be seen across experimental conditions in that {\sl (i)} the six color (C6) conditions performed better than the one color per object (C1) conditions and {\sl (ii)} color-overlap ($\neg$X) conditions performed better than object-exclusive color conditions (X) in the compositional generalization tests. This means that when we were training with one visible object at a time, the model would only generalize to pushing the yellow banana to the left if there were other yellow objects that had been push to the left. Color overlap benefitted the model to consider objects in the context of unseen actions.

\subsubsection{Showing two objects in the scene simultaneously, i.e. the object that is acted upon plus a distractor object, disrupted performance significantly} This can be seen in the middle two segments of Table~\ref{tab:results_sentence_all}. The difficulty of showing two objects at the same time is related to the binding problem (cf.~\cite{greff2020binding}) and comes down to the added challenge of not only having to classify one visible object, but having to decide which of the two objects that are visible is the correct option. This decision can only be made based on where the manipulator arm is moving, as the correct object is the one being acted upon. In this condition (V2), the best generalization performance is achieved in complex settings with 9 objects and 6 colors with around 16\% compositional generalization accuracy, while validation and test performances drop into the 30\% range. Hence, we can conjecture that solving this difficult case for the normal validation set might in turn benefit compositional generalization.

\subsubsection{Removing a sensory modality, i.e.~leaving away the joint readings, reduced training, validation and generalization performance} This can be seen in the bottom two segments of Table~\ref{tab:results_sentence_all}. From a technical point of view our models can be trained on the visual input alone, as the image data contains enough information to predict the interaction description. Therefore, we repeated the same experiments but trained without the joint input. This change results in a steep drop of the generalization performance, and noticeably impacts standard performance as well, giving clear evidence that multimodality is commonly beneficial in a scenario like ours. The impact was the largest for models that were shown four actions, which makes sense since the rotation of the joints is a clear giveaway of the action that is being performed. Without the joint positions the direction of the movements of either the manipulator or the object has to be extracted from the interpretation of the visual input in addition to the task of recognizing the object. We hypothesize that through the addition of different input channels, the performance of a model can be effectively increased with a potentially smaller impact on the network size and should be considered in the context of the particular machine learning task. We believe that in our experiments, because of multimodality, the model learns an implicit split of the action and the color/object that transfers into the representation in the network, boosting the generalization performance.
When we provided a rich enough target distribution in the setting with all objects and colors included, only missing two actions (i.e., V1-C6-O9-A2), generalization was better, at 50.28\%. This indicates that multimodality could be especially beneficial to generalization in cases with lower feature diversity.

Detailed analysis of the data, which is not shown in Table~\ref{tab:results_sentence_all}, let us also observe a significant difference in both the standard performance and compositional generalization performance between the four leave-out cases. The left-out interaction ``push left white football'' was not only recognized more often than the other leave-out cases when we trained with two visible objects and six colors per object, but both the standard performance and compositional generalization performance increased with the addition of more objects and actions to the training data. This could suggest that the model was not large enough to correctly classify most objects while also detecting the arm's interactions with the objects, but the football was easy enough to predict, in which case it comes close to the performance of the one-visible-objects case.

\section{Discussion}
A network that performs well on compositional generalization is expected to encode concepts of the world with compositional representations. We would expect that, e.g. actions, objects or colors are encoded as separable activation patterns. Whether and how such representations could be identified still needs to be investigated. In \cite{heinrich_2020}, the recurrent neural network's activations are visualized using PCA, showing that clusters emerged for the processing of actions, and to a lesser degree for objects and for colors. Our finding of varying degrees of the compositional generalization indicates that, if compositional representations exist in our network, these would be emerging rather gradually than in a clear-cut manner. 
In simple recurrent neural network models, such as the Hopfield network, the phenomenon of attractor states arises, which are a discrete set of activation patterns that the network states tend to converge to \cite{HertzKroghPalmer}. Such associative memory enforces robust processing of discrete entities, and the recurrent weights of our LSTM model could contribute to the network preferring certain activation patterns over others. Compositional generalisation moreover would presuppose separate discretization of representations for, e.g., actions, objects and verbs.

Conceptually, discrete compositional representations might be linked to minimal semantic units called sememes, which we deem language analogues of minimal memory units called engrams \cite{guan2016}. As it has been shown that the use of sememe knowledge bases can boost language tasks \cite{FanchaoQi2019,YujiaQin2020}, we suggest that compositional representations, once obtained, can be equally helpful.

\section{Summary}

We have shown the importance of a rich dataset for the emergence of compositional generalization.
On a single model architecture with unchanged parameters and on identical leave-out generalization test data, the generalization accuracy could be increased from 0\% to over 60\% by supplying additional training data to the models.
Better generalization accuracies were achieved in more complex settings, in which, as expected, lower accuracies were observed on the training and regular validation data.
It was most helpful for the generalization to add further objects, particularly when these had color overlaps with the leave-out data.
Adding two more actions led to an increase of generalization accuracy in many but not all conditions.
Adding a distractor object to the scene led to a large drop in generalization performance, and, in more complex settings, also of training and validation performance, suggesting an impact of the binding problem.
Removing the agent's proprioception by omitting its joint readings led to a drop in all, training, validation and generalization errors.
Our investigations are to the best of our knowledge the most systematic so far in a 3D simulated realistic robot environment.
While we demonstrated the importance of the data on compositional generalization, ongoing and future work on the influence of model architectures is another avenue of interest to the problem.

\section{Acknowledgment}
We thank Matthias Kerzel for providing the initial simulation setup and Ozan \"Ozdemir and Shengding Hu for valuable feedback on the document. The authors gratefully acknowledge support from the German Research Foundation DFG, project CML (TRR 169).

\bibliographystyle{unsrt}
\bibliography{references}

\end{document}